\documentclass{article}
\usepackage{spconf,amsmath,graphicx,hyperref}
\usepackage{tabularx, multirow, booktabs, comment}
\usepackage{tipa}
\usepackage{todonotes}

\usepackage{etoolbox}
\patchcmd{\thebibliography}
  {\settowidth}
  {\setlength{\itemsep}{0pt}\setlength{\parskip}{0pt}\settowidth}
  {}{}

\title{EZ-Ar-VC: Robust Arabic Voice Anonymization with Generalization across Dialects and Code-Switching}
\makeatletter
\gdef\@name{{\em Injy Hamed,$^{1}$ Adarsh Arigala,$^{2}$ Rufael Marew,$^{1}$ Arjun Gangwar,$^{2}$ Atharva Kulkarni,$^{1}$}\\
            {\em Hanan Aldarmaki,$^{1}$ Srinivasan Umesh$^{2}$}\\}
\makeatother
\address{$^1$Mohamed bin Zayed University of Artificial Intelligence, $^2$Indian Institute of Technology Madras}
\begin{document}
%
\maketitle
\begin{abstract}
Voice conversion (VC) models can be used to anonymize speech data while preserving naturalness. Existing models, however, are mainly optimized for English and other high-resource languages, and they fail to generalize well to Arabic due to its distinct phonology. In this work, we comprehensively evaluate existing zero-shot VC systems for Arabic speech, identifying limitations in pronunciation preservation. We then train an Arabic-centric model based on EZ-VC on diverse Arabic speech, and demonstrate strong performance in intelligibility, naturalness, and pronunciation preservation, while maintaining high speaker dissimilarity. Through subjective evaluation, we further demonstrate its robustness to code-switching and cross-dialectal conversion, as well as its generalization to Arabic dialects with limited representation in the training data. Our findings highlight the limitations of zero-shot VC and the importance of target-language adaptation for robust voice anonymization. 


\end{abstract}
\section{Introduction}
\label{sec:intro}


Speech synthesis systems have reached remarkable degrees of intelligibility and naturalness that are indistinguishable from human speech in the best cases. Voice Conversion (VC) systems are a subset of speech synthesis models that can convert a source utterance to sound like a target speaker, copying various speaker-specific characteristics. Typically, these models work best with enough representation of the target speaker, but recent zero-shot models have also shown competitive performance with only a short target utterance~\cite{your-tts,seed-vc,Diff-HierVC,joglekar-etal-2025-ez}. While applications of voice conversion could be used constructively for media and entertainment, there are several harmful use cases that could result from target voice imitation. 
On the other hand, the same VC technology could be utilized constructively as a means for Voice Anonymization (VA) to protect user identity. In our work, we are interested in anonymizing user data in speech datasets. Currently, large efforts in voice data collection, particularly in medical or educational applications, are kept proprietary due to privacy risks and the difficulty of perfectly protecting user identities without losing important features in the spoken utterances, such as accent and style.  While existing zero-shot models could be used for this purpose, their generalization and robustness have not been fully demonstrated across different languages. 

In this work, we rigorously evaluate 
VC models, particularly zero-shot models, for Arabic speech. Arabic is a pluricentric language with multiple regional dialects, spoken by over 300 million people as a first language ~\cite{ethnologue2024arabic}. 
While Modern Standard Arabic (MSA) is medium-resourced, most dialctal Arabic (DA) varieties  are low-resourced and underrepresented in speech and language technologies. 
Arabic speakers also 
code-switch with English and/or French, posing additional challenges. 
Our evaluation of existing VC models 
exposes various artifacts that limit their applicability for robust Arabic VA. For instance, the recently proposed EZ-VC model~\cite{joglekar-etal-2025-ez}, was shown to generalize  zero-shot across unseen languages. However, the 
evaluation was limited to Indic and European languages. Our evaluation on Arabic reveals that, while natural sounding, the converted speech is heavily accented. We conduct a phoneme-level analysis and find that this phenomenon is driven by sounds that are unique to  Arabic and likely underrepresented 
in the model’s training data, such as pharangeals, emphatics, and uvulars. 
Other VC models also show artifacts affecting intelligibility and naturalness.

These results underscore the importance of targeted efforts 
to preserve the 
characteristics of spoken Arabic varieties. 
While some Arabic-centric efforts exist
~\cite{toyin-etal-2024-sttatts,Arvoice}, these efforts treat VC as an auxiliary task and do not specifically optimize for 
conversion and anonymization. To close this gap, 
we introduce EZ-Ar-VC; a robust Arabic-centric VA model based on EZ-VC and demonstrate its robustness through objective and subjective evaluations. 
Our results 
show that the resulting speech is sufficiently anonymized while preserving 
intelligibility, naturalness, accent, and style. We also show that the model generalizes to 
code-switching with both English and French. EZ-Ar-VC is the first 
VC model 
shown to generalize robustly across Arabic varieties. The model is made publicly available as an open-source general-purpose Arabic VA tool to support privacy-preserving voice data sharing. We
provide our code, model checkpoint and demo
samples here: \url{https://github.com/EZ-AR-VC/EZ-AR-VC}

\section{Related Work}
Recent VC research has increasingly shifted toward any-to-any systems that can generalize to unseen speakers. 
Approaches include discrete-unit flow matching, such as EZ-VC~\cite{joglekar-etal-2025-ez}, end-to-end TTS-VC hybrids such as YourTTS~\cite{your-tts}, diffusion-based models such as Diff-HierVC~\cite{Diff-HierVC} and Seed-VC~\cite{seed-vc}, and SSL retrieval/prompted vocoding, such as kNN-VC~\cite{baas23_interspeech} and Vec2wav~2.0~\cite{guo2024vec2wav}. 
However, evaluations 
have predominantly focused on English, European, and Indic languages. 
While limited research in VA has addressed multilingual and code-switching scenarios~\cite{meyer2024multilingual,meyer2025codeswitch}, 
Arabic still remains largely underexplored. 
ArVoice~\cite{Arvoice}, a multi-speaker MSA corpus with seven human and four synthetic voices, demonstrates VC, 
but its predominantly read MSA speech provides limited conversational, dialectal, and code-switched coverage. Abdullah et al ~\cite{abdullah2025voice} used kNN-VC to augment dialect-identification data; 
however, their evaluation targets dialect identification rather than conversion quality. Our work fills this gap by benchmarking and adapting VC models to Arabic and evaluating their robustness across dialectal and code-switching variants. 
We provide a comprehensive objective and subjective evaluation of 
speaker privacy, 
intelligibility, pronunciation, naturalness, and style preservation.

\section{Methodology}

\subsection{Datasets}
\label{sec:datsets}
We construct our training and evaluation datasets to span MSA, dialectal Arabic, and code-switched speech. 
We utilize \textbf{unlabeled data} for training. For MSA, we use QASR~\cite{mubarak-etal-2021-qasr} (2K hrs). For dialectal Arabic, we use TunSwitch~\cite{Tunswitch} (153 hours), covering Tunisian Arabic with instances of code-switching with French, and  ADI17~\cite{ADI17} ($\sim$3K hrs), covering speech from 17 Arabic-speaking countries. 
For foreign languages, we use 550 hours of LibriTTS English~\cite{zen2019libritts} and 500 hours of French Common Voice speech~\cite{commonvoice-2020}.

We also utilize \textbf{labeled data} 
to construct source-speaker development and test sets and target-speaker set, while allocating a subset of the data for training.  For MSA, we use ArVoice ~\cite{Arvoice} and MASC~\cite{MASC}. 
For dialectal Arabic, we use Casablanca~\cite{casablanca}, covering 
Algeria, Egypt, Jordan, Mauritania, Morocco, Palestine, UAE, and Yemen.  We also use a small proprietary in-house dataset of Emirati Arabic speech for evaluation, referred to as RamsaLab. 
For code-switching, we use ArzEn~\cite{HVA20} for Egyptian-English, Mixat~\cite{al2024mixat} for Emirati-English, and TEDxTN~\cite{bougares2025tedxtn} for Tunisian-French. 
We ensure no overlap between source and target speakers.\footnote{Given current limitations in Arabic ASR, we restrict the source-speaker development and test sets to utterances with a CER below 5\%, enabling a more reliable intelligibility evaluation as outlined in Section~\ref{sec:eval_metrics}} 
Table~\ref{tab:corpus_stats} summarizes the labeled corpora statistics across partitions.

\begin{table}[]
\centering
\caption{Statistics of the labeled corpora used for 
training, source-speaker development and test, and target-speaker partitions, reporting duration in hours (H) and 
speaker count (S).}
\label{tab:corpus_stats}
\setlength{\tabcolsep}{3.5pt}
\resizebox{0.9\columnwidth}{!}{
\begin{tabular}{lcccccccc}
\toprule
 & \multicolumn{2}{c}{\textbf{Train}} & \multicolumn{2}{c}{\textbf{Source\ Dev}} & \multicolumn{2}{c}{\textbf{Source\ Test}} & \multicolumn{2}{c}{\textbf{Target}} \\\cmidrule(lr){2-3}\cmidrule(lr){4-5}\cmidrule(lr){6-7}\cmidrule(lr){8-9}
 & \multicolumn{1}{c}{\textbf{H}} & \multicolumn{1}{c}{\textbf{S}} & \multicolumn{1}{c}{\textbf{H}} & \multicolumn{1}{c}{\textbf{S}} & \multicolumn{1}{c}{\textbf{H}} & \multicolumn{1}{c}{\textbf{S}} & \multicolumn{1}{c}{\textbf{H}} & \multicolumn{1}{c}{\textbf{S}} \\\midrule
\textbf{ArVoice} & 3.9 & 2 & 1.4 & 1 & 2.2 & 2 & 31.0 & 6 \\
\textbf{MASC} & 485.0 & 0 & 6.6 & 101 & 6.0 & 90 & 0.0 & 0 \\
\textbf{Casablanca} & 0.0 & 0 & 1.5 & - & 1.5 & - & 0.0 & 0 \\
\textbf{ArzEn} & 5.6 & 14 & 2.1 & 11 & 2.5 & 13 & 0.4 & 2 \\
\textbf{Mixat} & 3.2 & 6 & 0.2 & 4 & 0.2 & 4 & 5.9 & 1 \\
\textbf{TEDxTN} & 22.3 & 96 & 0.3 & 3 & 0.6 & 6 & 0.4 & 2 \\
\textbf{RamsaLab} & 0.0 & 0 & 0.1 & 16 & 0.1 & 17 & 0.0 & 0 \\
\bottomrule
\end{tabular}}
\end{table}

\subsection{Voice Conversion Models}
We benchmark six zero-shot VC models:  Diff-HierVC~\cite{Diff-HierVC}, Seed-VC~\cite{seed-vc}, EZ-VC~\cite{joglekar-etal-2025-ez}, kNN-VC~\cite{baas23_interspeech}, YourTTS~\cite{your-tts}, and Vec2Wav 2.0~\cite{guo2024vec2wav}.  Additionally, we train Arabic-centric models based on EZ-VC and kNN-VC (the best two baselines from preliminary listening tests). For each architecture, we train two variants: (1) \textbf{(Ar)}, trained on $\sim$5K hrs of Arabic data; and (2) \textbf{(AR+EN+FR)}, trained on $\sim$6K hrs of multilingual data (English and French in addition to Arabic),  as described in Section~\ref{sec:datsets}. 

\textbf{EZ-VC:} We build a diffusion-based unit-to-speech model following the EZ-VC approach~\cite{joglekar-etal-2025-ez}. Discrete speech units are extracted from the 14th layer of XEUS~\cite{chen-etal-2024-towards-robust}, which operates at 16 kHz. For the EZ-VC(AR) setting, 
we train a 2K-cluster k-means model on 300 hours of Arabic speech. For the EZ-VC(AR+EN+FR) setting, 
we train a separate 2K-cluster model on the same Arabic data in addition to 50 hours each of English and French. In both settings, we use F5-TTS~\cite{chen-etal-2025-f5} as the backbone and extend its vocabulary with the corresponding discrete units, initializing their embeddings by randomly sampling from the existing F5-TTS vocabulary to facilitate faster convergence. 
Starting from the F5-TTS model pretrained on 100K hours of speech, we fine-tune the two unit-to-speech models for 150K steps. 
\begin{table*}[t]
\centering
\caption{Results for objective evaluation. 
Intelligibility is reported using WER and CER (\%). 
SIM$_{\text{src}}$ and SIM$_{\text{tgt}}$ denote speaker
similarity of the converted speech to the source and target speaker,
respectively. ASV EER (\%) is the equal error rate of an automatic speaker
verification system on the converted speech. Pronunciation distance is measured as JSD between phonetic posteriorgrams, across emphatics (E), uvulars (U), pharyngeals (P), vowels (V), and all phonemes.}
\label{tab:objective_eval_all}
\footnotesize
\setlength{\tabcolsep}{4pt}
\begin{tabular}{lcccccccccccc}
\toprule
\multirow{2}{*}{Model} & \multicolumn{2}{c}{Intelligibility} & \multicolumn{3}{c}{Privacy} 
& Naturalness & \multicolumn{5}{c}{Pronunciation Distance}\\
\cmidrule(lr){2-3} \cmidrule(lr){4-6} 
\cmidrule(lr){7-7} \cmidrule(lr){8-12}
 & WER $\downarrow$ & CER $\downarrow$ & SIM$_{\text{src}}$ $\downarrow$ & SIM$_{\text{tgt}}$ $\uparrow$ & EER $\uparrow$ & UTMOS $\uparrow$ & All $\downarrow$ & E $\downarrow$ & U $\downarrow$ & P $\downarrow$ & V $\downarrow$\\
\midrule
Vec2Wav 2.0               & 36.0 & 21.5 & \textbf{0.21} & 0.52 & 46.28 & 2.95 & 0.24 & 0.58 & 0.55 & 0.24 & 0.22 \\
YourTTS               & 35.2 & 20.5 & 0.22 & 0.29 & 35.88 & 2.03 & 0.29 & 0.58 & 0.39 & 0.33 & 0.26 \\
Diff-HierVC           & 23.2 & 12.0 & 0.29 & 0.33 & 42.65 & 2.54 & 0.24 & 0.44 & 0.37 & 0.23 & 0.21 \\
Seed-VC               & 17.7 & 7.7  & 0.22 & 0.57 & \textbf{49.46} & 2.71 & 0.15 & 0.44 & 0.34 & 0.20 & 0.16 \\
\midrule
kNN-VC (baseline)    & 21.1 & 10.2 & 0.32 & 0.55 & 45.95 & 2.75 & 0.14 & 0.34 & 0.22 & 0.14 & \textbf{0.12} \\
kNN-VC (AR)           & 20.8 & 10.1 & 0.31 & 0.55 & 46.72 & 2.76 & 0.14 & 0.32 & 0.22 & 0.14 & \textbf{0.12} \\
kNN-VC (AR+EN+FR)     & 20.4 & 9.7  & 0.30 & 0.56 & 46.32 & 2.75 & 0.14 & 0.31 & 0.21 & \textbf{0.13} & \textbf{0.12} \\
\midrule
EZ-VC (baseline)     & 30.0 & 14.9 & 0.22 & \textbf{0.59} & 45.60 & \textbf{3.11} & 0.18 & 0.57 & 0.54 & 0.42 & 0.19 \\
EZ-VC (AR)            & \textbf{16.7} & \textbf{7.1} & 0.31 & 0.55 & 43.87 & 2.71 & \textbf{0.12} & \textbf{0.27} & \textbf{0.14} & \textbf{0.13} & 0.14 \\
EZ-VC (AR+EN+FR)      & 16.8 & \textbf{7.1} & 0.31 & 0.57 & 44.54 & 2.72 & \textbf{0.12} & \textbf{0.27} & \textbf{0.14} & \textbf{0.13} & 0.14 \\
\bottomrule
\end{tabular}
\end{table*}

\textbf{KNN-VC:} We implement the 
kNN-VC model~\cite{baas23_interspeech}, whichperforms non-parametric voice conversion, retrieving acoustically similar frames from the target speaker's speech. We extract frame-level representations from the 6th layer of a pretrained WavLM model 
and construct a target speaker feature database from the available utterances. For each source speech frame, the model retrieves the $k=4$ nearest target utterance feature neighbors and uses their representation to construct a converted feature sequence. We train a multi-speaker HiFi-GAN vocoder to synthesise waveforms from the converted WavLM features. We follow the original kNN-VC vocoder architecture and training configuration. 
All audio is resampled to 16~kHz for WavLM feature extraction and vocoder training.

\subsection{Evaluation Metrics}
\label{sec:eval_metrics}
We evaluate \textbf{privacy} using speaker dissimilarity and Equal Error Rate (EER). For speaker dissimilarity, we extract speaker embeddings using a WavLM-L-based 
ECAPA-TDNN model and compute cosine similarity between generated and source/target audio. For EER, we constructed $\sim$170K balanced ASV trials from 123 speakers across all datasets, reserving one original utterance per speaker for enrollment and excluding it from testing. Each converted test recording was paired with same-speaker and different-speaker enrollment recordings from the same dataset. 
We use the VoicePrivacy Challenge 2024 toolkit’s\footnote{\scriptsize\url{https://github.com/Voice-Privacy-Challenge/Voice-Privacy-Challenge-2024}} pretrained ECAPA-TDNN model, cosine similarity scoring, and SpeechBrain’s EER routine.

For utility, we assess \textbf{intelligibility}, with Word Error Rate (WER) and Character Error Rate (CER) from an ASR model\footnote{\scriptsize\url{https://huggingface.co/oddadmix/Whisperv3-tunisian-codeswitch}} specialized in code-switched Tunisian-French for TEDxTN and Cohere Arabic ASR model~\footnote{\scriptsize\url{https://huggingface.co/CohereLabs/cohere-transcribe-arabic-07-2026}} 
for the remaining corpora as it is trained on Arabic and English. 
We evaluate \textbf{naturalness } using UTMOS~\cite{UTMOS}. 
We also assess \textbf{accentedness}, measuring the extent to which generated audio exhibits native-like pronunciation. 
Following~\cite{zhong2025ppgs}, we use phonetic posteriorgrams (PPGs) to evaluate pronunciation distance. We extract PPGs from source and generated audio using a phoneme recognizer~\cite{xu2021simple}, align them using Dynamic Time Warping, and 
compute Jensen-Shannon divergence (JSD) between aligned pairs. 
We also report JSD across emphatic, uvular, and pharyngeal classes, which are challenging Arabic phonetic structures
, and for vowels, motivated by vowel-duration deviations observed in our internal evaluations.\footnote{Emphatics are only represented by \textipa{/D/}, as 
/d\super{\textrevglotstop}/, /s\textsuperscript{\textrevglotstop}/, and /t\textsuperscript{\textrevglotstop}/
are not recognized by the phonemizer.}

\subsection{Human Evaluation Setup}
We conduct two human evaluation studies with the same ten annotators recruited through Prolific.\footnote{
Each annotator received a compensation of \$75, based on an hourly rate of \$12 with an average of 6.25 hours of annotation work.
} The annotators are gender-balanced, aged 22-37 ($\mu=29$, $\sigma=5$), and represent diverse nationalities, including Egyptian, Moroccan, Palestinian, Saudi Arabian, Tunisian, and Emirati.

\textbf{Human Study 1:} The first study compares the performance of six VC systems: Vec2Wav 2.0, Seed-VC, kNN-VC baseline, kNN-VC (AR+EN+FR), EZ-VC baseline, and EZ-VC (AR+EN+FR). We sample 20 monolingual Arabic utterances evenly from ArVoice, ArzEn, Mixat, and TEDxTN, covering MSA, Egyptian, Emirati, and Tunisian Arabic, with target speakers selected from the same corpus. Each annotator rated all 120 generated samples on six dimensions: speaker dissimilarity (between source and generated audio), intelligibility, naturalness, accentedness, style preservation, and audio quality, 
using five-point scales with predefined rubrics.

\textbf{Human Study 2:} We further evaluate the robustness of our best-performing model along three directions: code-switching, cross-dialectal conversion, and generalization to limited-resource dialects. In addition to the 20 previous utterances, we extend the evaluation by: (a) adding 15 code-switched utterances evenly sampled from ArzEn, Mixat, and TEDxTN, with same-corpus target speakers; (b) pairing the 35 monolingual and code-switched utterances considered thus far with target speakers of different Arabic varieties; and (c) adding 15 utterances evenly sampled across Yemeni, Algerian, and Palestinian Arabic, representing dialects with limited training data. Collectively, the study comprises 85 source-target pairs, that provide a balanced representation of gender and language variants across source and target speakers. 
Each sample is evaluated by all the annotators for the same dimensions as in Human Study 1.
\section{Evaluation and Analysis}
\label{sec:eval}


\subsection{Performance Across Models}
We report the objective evaluation results in Table~\ref{tab:objective_eval_all} and the subjective evaluation results in Table~\ref{tab:subjective_eval_1}. 
Overall, the baselines exhibit trade-offs between privacy and utility across the evaluated systems. YourTTS performs poorly across most dimensions, while Diff-HierVC shows no clear strength in either privacy or utility, and thus were not included in our human evaluation study. Vec2Wav 2.0 and EZ-VC both achieve strong privacy, but at the expense of intelligibility and Arabic pronunciation. Seed-VC provides better performance, demonstrating strong privacy and intelligibility, although challenges remain in the pronunciation of Arabic-specific phonemes. Conversely, kNN-VC provides the strongest baseline for Arabic pronunciation, but is less effective than Seed-VC in terms of privacy. 
For the trained Arabic-centric models, kNN-VC shows only modest improvements over its baseline. In contrast, significant gains are observed for EZ-VC, with EZ-VC (AR+EN+FR) performing best overall on intelligibility and Arabic pronunciation while maintaining effective speaker conversion (SIM$_{\text{tgt}}$) and privacy (EER).


The strength of EZ-VC(AR+EN+FR) is further demonstrated by the human evaluation, where it ranks first in intelligibility, naturalness, accentedness, style preservation, and audio quality, while maintaining high speaker dissimilarity (4.02). According to our annotation rubric, this corresponds to listeners perceiving the source and generated speakers as `very different’, indicating strong privacy. Given the performance of EZ-VC(AR+EN+FR) across evaluation dimensions, 
we identify it as our best-performing model and refer to it as \textbf{EZ-Ar-VC}.
\vspace{-0.2cm}
\subsection{Robustness Evaluation of EZ-Ar-VC}
Table~\ref{tab:subjective_eval_2} reports the results of Human Study 2, further evaluating EZ-Ar-VC under three conditions: (a) monolingual versus code-switched source speech; (b) different Arabic varieties for source$\rightarrow$target conversion; and (c) additional Yemeni, Algerian, and Palestinian Arabic 
source dialects. Results demonstrate that EZ-Ar-VC remains robust across the challenging conditions of code-switching, cross-dialectal conversion, and low-resource Arabic dialects. 
Intelligibility remains consistently high (4.62-4.95), regardless of code-switching dimension or source and target Arabic variant mismatch. Naturalness is similarly robust. 
Speaker dissimilarity remains above 4 in most cases, except for TEDxTN and ArzEn, 
where slight degradation is observed, which could be due to greater similarity between source and target speakers in the sampled data. 
The model also generalizes well to the low-resource Algerian, Palestinian, and Yemeni Arabic, maintaining intelligibility above 4.6 and scores above 4 for most other dimensions. Overall, these results demonstrate the robustness of the model across diverse linguistic conditions. 

\begin{table}[t]
\centering
\caption{Results for Human Study 1 showing MOS for intelligibility (I), speaker dissimilarity (S), naturalness (N), accentedness (A), style preservation (P), and audio quality (Q).}
\label{tab:subjective_eval_1}
\setlength{\tabcolsep}{2.5pt}
\resizebox{0.8\columnwidth}{!}{
\begin{tabular}{lrrrrrr}
\toprule
\multicolumn{1}{l}{Models} & \multicolumn{1}{c}{I $\uparrow$} & \multicolumn{1}{c}{S $\uparrow$} & \multicolumn{1}{c}{N $\uparrow$} & \multicolumn{1}{c}{A $\uparrow$} & \multicolumn{1}{c}{P $\uparrow$} & \multicolumn{1}{c}{Q $\uparrow$} \\ \midrule
Vec2Wav 2.0 & 4.38 & 4.15 & 3.95 & 4.08 & 4.52 & 4.12 \\
Seed-VC & 4.70 & 4.18 & 4.09 & 4.27 & 4.64 & 4.35 \\
kNN-VC (baseline) & 4.81 & 3.96 & 4.39 & 4.62 & 4.69 & 4.51 \\
kNN-VC (AR+EN+FR) & 4.82 & 3.93 & 4.43 & 4.65 & 4.63 & 4.49 \\
EZ-VC (baseline) & 4.34 & \textbf{4.19} & 3.68 & 3.81 & 4.27 & 4.04\\
EZ-VC (AR+EN+FR) & \textbf{4.84} & 4.02 & \textbf{4.53} & \textbf{4.67} & \textbf{4.77} & \textbf{4.55} \\\hline
\end{tabular}}
\end{table}

\vspace{-0.2cm}\begin{table}[t]
\centering
\caption{Results for Human Study 2 showing MOS for intelligibility (I), speaker dissimilarity (S), naturalness (N), accentedness (A), style preservation (P), and audio quality (Q).
DA denotes dialectal Arabic, and DA$_x$$\rightarrow$ DA$_x$/DA$_x$$\rightarrow$ DA$_y$ indicate same/different source$\rightarrow$target dialects, respectively.}
\label{tab:subjective_eval_2}
\setlength{\tabcolsep}{2.5pt}
\resizebox{0.75\columnwidth}{!}{
\begin{tabular}{lrrrrrr}
\toprule
\multicolumn{1}{l}{Models} & \multicolumn{1}{c}{I $\uparrow$} & \multicolumn{1}{c}{S $\uparrow$} & \multicolumn{1}{c}{N $\uparrow$} & \multicolumn{1}{c}{A $\uparrow$} & \multicolumn{1}{c}{P $\uparrow$} & \multicolumn{1}{c}{Q $\uparrow$} \\ \midrule
\multicolumn{7}{c}{(a) Source Monolingual vs. Code-switching (CS)}\\\hline
Egyptian$_{mono}$ & 4.95 & 3.75 & 4.50 & 4.80 & 4.80 & 4.69 \\
Egyptian$_{CS}$ & 4.87 & 3.84 & 4.52 & 4.70 & 4.67 & 4.61 \\
Tunisian$_{mono}$ & 4.86 & 3.53 & 4.42 & 4.66 & 4.72 & 4.51 \\
Tunisian$_{CS}$ & 4.82 & 3.76 & 4.47 & 4.51 & 4.65 & 4.56 \\
Emirati$_{mono}$ & 4.93 & 4.86 & 4.59 & 4.75 & 4.69 & 4.66 \\
Emirati$_{CS}$ & 4.86 & 4.79 & 4.64 & 4.74 & 4.78 & 4.70 \\\hline
\multicolumn{7}{c}{(b) Source-Target Speaker Dialectal Variants}\\\hline
MSA$\rightarrow$MSA & 4.82 & 4.46 & 4.58 & 4.44 & 4.84 & 4.76 \\
MSA$\rightarrow$DA & 4.86 & 4.36 & 4.58 & 4.50 & 4.66 & 4.82 \\
DA$\rightarrow$MSA & 4.91 & 4.50 & 4.28 & 4.75 & 4.71 & 4.66 \\
DA$_{x}$$\rightarrow$DA$_{x}$ & 4.86  & 3.97 & 4.55 & 4.71 & 4.74 & 4.60 \\
DA$_{x}$$\rightarrow$DA$_{y}$ & 4.90 & 4.07 & 4.60 & 4.65 & 4.70 & 4.64 \\\hline
\multicolumn{7}{c}{(c) Source Dialectal Variation}\\\hline
MSA & 4.84 & 4.41 & 4.58 & 4.47 & 4.75 & 4.79 \\
Algerian & 4.74 & 4.32 & 4.34 & 4.44 & 4.52 & 4.62 \\
Palestinian & 4.80 & 4.16 & 3.86 & 4.26 & 4.58 & 4.64 \\
Yemeni & 4.62 & 4.34 & 4.16 & 4.26 & 4.38 & 4.42 \\\bottomrule 
\end{tabular}}
\end{table}

\section{Conclusion}

In this work, we demonstrated limitations of various open-source voice conversion models and their failure to generalize to spoken Arabic varieties. We analyzed the cause of this failure and revealed patterns that relate to the lack of representation of target language phones, leading to unnatural sounding accented speech. Our contribution lies in revealing these limitations of zero-shot voice conversion, 
as well as offering an open-source alternative model for Arabic voice anonymization that is trained with diverse Arabic data and shown to  generalize robustly across 
different dialects. Through multi-dimensional subjective analysis, the model is shown to preserve intelligibility, naturalness, accent, and style, including for variants with limited resources and instances of cross-lingual code-switching. 
\newpage
\vspace{-0.4cm}
\bibliographystyle{IEEEbib}
\bibliography{main}

@inproceedings{joglekar-etal-2025-ez,
    title = "{EZ}-{VC}: Easy Zero-shot Any-to-Any Voice Conversion",
    author = "Joglekar, Advait  and
      Singh, Divyanshu  and
      Bhatia, Rooshil Rohit  and
      Umesh, Srinivasan",
    booktitle = "EMNLP",
    year = "2025",
    url = "https://aclanthology.org/2025.findings-emnlp.1077/",
    doi = "10.18653/v1/2025.findings-emnlp.1077",
    ISBN = "979-8-89176-335-7",
    
}

@inproceedings{chen-etal-2025-f5,
    title = "F5-{TTS}: A Fairytaler that Fakes Fluent and Faithful Speech with Flow Matching",
    author = "Chen, Yushen  and
      Niu, Zhikang  and
      Ma, Ziyang  and
      Deng, Keqi  and
      Wang, Chunhui  and
      Zhao, Jian  and
      Yu, Kai  and
      Chen, Xie",
    booktitle = "Proceedings of ACL",
    year = "2025",
    url = "https://aclanthology.org/2025.acl-long.313/",
    doi = "10.18653/v1/2025.acl-long.313",

    ISBN = "979-8-89176-251-0",
   
}

@inproceedings{chen-etal-2024-towards-robust,
    title = "Towards Robust Speech Representation Learning for Thousands of Languages",
    author = "Chen, William  and
      Zhang, Wangyou  and
      Peng, Yifan  and
      Li, Xinjian  and
      Tian, Jinchuan  and
      Shi, Jiatong  and
      Chang, Xuankai  and
      Maiti, Soumi  and
      Livescu, Karen  and
      Watanabe, Shinji",
    booktitle = "Proceedings of the EMNLP",
    year = "2024",
    url = "https://aclanthology.org/2024.emnlp-main.570/",
    doi = "10.18653/v1/2024.emnlp-main.570",
}

@inproceedings{commonvoice-2020,
  author = {Ardila, R. and Branson, M. and Davis, K. and Henretty, M. and Kohler, M. and Meyer, J. and Morais, R. and Saunders, L. and Tyers, F. M. and Weber, G.},
  title = {{Common Voice: A Massively-Multilingual Speech Corpus}},
  booktitle = {LREC},
  year = 2020
}

@inproceedings{zen2019libritts,
    author    = {Heiga Zen and Viet Dang and Rob Clark and Yu Zhang and Ron J. Weiss and Ye Jia and Zhifeng Chen and Yonghui Wu},
    title     = {{LibriTTS: A Corpus Derived from LibriSpeech for Text-to-Speech}},
    booktitle = {Proceedings of Interspeech},
    doi       = {10.21437/Interspeech.2019-2441},
    year      = {2019},
}

@inproceedings{zhong2025ppgs,
    author={Zhong, J. and Liu, S. and Wells, D. and Richmond, K.},
    title={Pairwise evaluation of accent similarity in speech synthesis},
    booktitle={Proceedings of Interspeech},
    year={2025},
}

@inproceedings{baas23_interspeech,
  title     = {{Voice Conversion With Just Nearest Neighbors}},
  author    = {Matthew Baas and Benjamin {van Niekerk} and Herman Kamper},
  year      = {2023},
  booktitle = {Proceedings of Interspeech},
  doi       = {10.21437/Interspeech.2023-419},
  issn      = {2958-1796},
}

@inproceedings{UTMOS,
    title     = {{UTMOS: UTokyo-SaruLab System for VoiceMOS Challenge}},
    author    = {Takaaki Saeki and Detai Xin and Wataru Nakata and Tomoki Koriyama and Shinnosuke Takamichi and Hiroshi Saruwatari},
    booktitle = {Proceedings of Interspeech},
    doi       = {10.21437/Interspeech.2022-439},
    year      = {2022},
}

@inproceedings{HVA20,
  title={{ArzEn}: A Speech Corpus for Code-switched {E}gyptian {A}rabic-{E}nglish},
  author={Hamed, Injy and Vu, Ngoc Thang and Abdennadher, Slim},
  booktitle={Proceedings of LREC},
  year={2020}
}

@inproceedings{al2024mixat,
  title={Mixat: A Data Set of Bilingual {E}mirati-{E}nglish Speech},
  author={Al Ali, Maryam Khalifa and Aldarmaki, Hanan},
  booktitle={Proceedings of SIGUL},
  year={2024}
}

@inproceedings{Arvoice,
  title     = {{ArVoice: A Multi-Speaker Dataset for Arabic Speech Synthesis}},
  author    = {Hawau Toyin and Rufael Marew and Humaid Alblooshi and Samar M. Magdy and Hanan Aldarmaki},
  year      = {2025},
  booktitle = {{Proceedings of Interspeech}},
  pages     = {4808--4812},
  doi       = {10.21437/Interspeech.2025-1550},
  issn      = {2958-1796},
}

@inproceedings{MASC,
  author={Al-Fetyani, Mohammad and Al-Barham, Muhammad and Abandah, Gheith and Alsharkawi, Adham and Dawas, Maha},
  booktitle={Proceedings of SLT}, 
  title={{MASC}: Massive {A}rabic Speech Corpus}, 
  year={2023},
  volume={},
  number={},
  pages={1006-1013},
  doi={10.1109/SLT54892.2023.10022652}}

@inproceedings{mubarak-etal-2021-qasr,
    title = "{QASR}: {QCRI} Aljazeera Speech Resource A Large Scale Annotated {A}rabic Speech Corpus",
    author = "Mubarak, Hamdy  and
      Hussein, Amir  and
      Chowdhury, Shammur Absar  and
      Ali, Ahmed",
    booktitle = {Proceedings of ACL-IJCNLP},
    year = "2021", 
    url = "https://aclanthology.org/2021.acl-long.177/",
    doi = "10.18653/v1/2021.acl-long.177",
}

@inproceedings{ADI17,
  author={Shon, Suwon and Ali, Ahmed and Samih, Younes and Mubarak, Hamdy and Glass, James},
  booktitle={ICASSP}, 
  title={{ADI17}: A Fine-Grained {A}rabic Dialect Identification Dataset}, 
  year={2020},
  doi={10.1109/ICASSP40776.2020.9052982}}

@INPROCEEDINGS{Tunswitch,
  author={Abdallah, Ahmed Amine Ben and Kabboudi, Ata and Kanoun, Amir and Zaiem, Salah},
  booktitle={ICASSP}, 
  title={Leveraging Data Collection and Unsupervised Learning for Code-Switched Tunisian Arabic Automatic Speech Recognition}, 
  year={2024},
  volume={},
  number={},
  doi={10.1109/ICASSP48485.2024.10445734}}

@inproceedings{casablanca,
  title={Casablanca: Data and models for multidialectal arabic speech recognition},
  author={Talafha, Bashar and Kadaoui, Karima and Magdy, Samar Mohamed and Habiboullah, Mariem and Chafei, Chafei Mohamed and El-Shangiti, Ahmed Oumar and Zayed, Hiba and Tourad, Mohamedou Cheikh and Alhamouri, Rahaf and Assi, Rwaa and others},
  booktitle={Proceedings of EMNLP},
  year={2024}
}

@article{guo2024vec2wav,
  title={vec2wav 2.0: Advancing Voice Conversion via Discrete Token Vocoders},
  author={Guo, Yiwei and Li, Zhihan and Li, Junjie and Du, Chenpeng and Wang, Hankun and Wang, Shuai and Chen, Xie and Yu, Kai},
  journal={arXiv preprint arXiv:2409.01995},
  year={2024}
}

@inproceedings{Diff-HierVC,
  title     = {{Diff-HierVC: Diffusion-based Hierarchical Voice Conversion with Robust Pitch Generation and Masked Prior for Zero-shot Speaker Adaptation}},
  author    = {Ha-Yeong Choi and Sang-Hoon Lee and Seong-Whan Lee},
  year      = {2023},
  booktitle = {Proceedings of Interspeech},
  pages     = {2283--2287},
  doi       = {10.21437/Interspeech.2023-817},
  issn      = {2958-1796},
}

@InProceedings{your-tts,
  title = 	 {{Y}our{TTS}: Towards Zero-Shot Multi-Speaker {TTS} and Zero-Shot Voice Conversion for Everyone},
  author =       {Casanova, Edresson and Weber, Julian and Shulby, Christopher D and Junior, Arnaldo Candido and G{\"o}lge, Eren and Ponti, Moacir A},
  booktitle = {Proceedings of ICML},
  year = 	 {2022},
  pages = 	 {2709--2720},  
}

@article{seed-vc,
  title={Zero-shot voice conversion with diffusion transformers},
  author={Liu, Songting},
  journal={preprint arXiv:2411.09943},
  year={2024}
}

@article{xu2021simple,
  title={Simple and effective zero-shot cross-lingual phoneme recognition},
  author={Xu, Qiantong and Baevski, Alexei and Auli, Michael},
  journal={arXiv preprint arXiv:2109.11680},
  year={2021}
}

@inproceedings{toyin-etal-2024-sttatts,
    title = "{STTATTS}: Unified Speech-To-Text And Text-To-Speech Model",
    author = "Toyin, Hawau Olamide  and
      Li, Hao  and
      Aldarmaki, Hanan",
    booktitle = {Findings of EMNLP},
    year = "2024",
    url = "https://aclanthology.org/2024.findings-emnlp.401/",
    doi = "10.18653/v1/2024.findings-emnlp.401",
}

@inproceedings{abdullah2025voice,
  title     = {{Voice Conversion Improves Cross-Domain Robustness  for Spoken Arabic Dialect Identification}},
  author    = {Badr M. Abdullah and Matthew Baas and Bernd Möbius and Dietrich Klakow},
  year      = {2025},
  booktitle = {Proceedings of Interspeech},
  doi       = {10.21437/Interspeech.2025-1809},
  issn      = {2958-1796},
}

@inproceedings{meyer2024multilingual,
  author    = {Meyer, Sarina and Lux, Florian and Vu, Ngoc Thang},
  title     = {Probing the Feasibility of Multilingual Speaker Anonymization},
  booktitle = {Proceedings of Interspeech},
  year      = {2024},
  url       = {https://www.isca-archive.org/interspeech_2024/meyer24_interspeech.html}
}

@inproceedings{meyer2025codeswitch,
  author    = {Meyer, Sarina and Kolos, Ekaterina and Vu, Ngoc Thang},
  title     = {First Steps Towards Voice Anonymization for Code-Switching Speech},
  booktitle = {Proceedings of Interspeech},
  year      = {2025},
  doi       = {10.21437/Interspeech.2025-1027}
}

@misc{ethnologue2024arabic,
  author = {Eberhard, David M. and Simons, Gary F. and Fennig, Charles D.},
  title = {Ethnologue: Languages of the World},
  edition = {28th},
  year = {2025},
  publisher = {SIL International},
  address = {Dallas, Texas},
}

@inproceedings{bougares2025tedxtn,
  title={{TEDxTN}: A Three-way Speech Translation Corpus for Code-Switched {T}unisian {A}rabic-{E}nglish},
  author={Bougares, Fethi and Mdhaffar, Salima and Elleuch, Haroun and Est{\`e}ve, Yannick},
  booktitle={Proceedings of ArabicNLP},
  year={2025}
}

\end{document}